\documentclass{article}
\usepackage{ijcai21}

\usepackage{times}

\usepackage{soul}
\usepackage{url}
\usepackage[hidelinks]{hyperref}
\usepackage[utf8]{inputenc}
\usepackage[small]{caption}
\usepackage{graphicx}
\usepackage{amsmath}
\usepackage{booktabs}
\usepackage{stfloats}
\usepackage{algorithm}
\usepackage{algorithmic}
\usepackage{amssymb}
\usepackage{xcolor}
\usepackage{mathrsfs}
\usepackage{multirow}
\usepackage{multicol}

\newcommand{\etal}{\textit{et al}. }

\newcommand{\eg}{\emph{e.g.}}
\newcommand{\lrw}{\textbf{LRW}}

\newcommand{\headmotionpredictor}{\mathbf{N_{H}}}
\newcommand{\keypointgenerator} {\mathbf{N_{M}}} 
\newcommand{\keypointdetector}{\mathbf{N_{D}}}
\newcommand{\imagegenerator} {\mathbf{N_{I}}}

\title{Audio2Head: Audio-driven One-shot Talking-head Generation\\ with Natural Head Motion}

\author{
Suzhen Wang$^1$\and
Lincheng Li$^1$\and
Yu Ding$^1$\thanks{Corresponding author.}\and
Changjie Fan$^1$\and
Xin Yu$^2$
\affiliations
$^1$ Virtual Human Group, Netease Fuxi AI Lab, China\\
$^2$University of Technology Sydney
\emails
\{wangsuzhen, lilincheng, dingyu01, fanchangjie\}@corp.netease.com,
xin.yu@uts.edu.au
}

\begin{document}

\maketitle

\begin{abstract}

We propose an audio-driven talking-head method to generate photo-realistic talking-head videos from a single reference image. In this work, we tackle two key challenges: \textit{(i)} producing natural head motions that match speech prosody, and \textit{(ii)} maintaining the appearance of a speaker in a large head motion while stabilizing the non-face regions. We first design a head pose predictor by modeling rigid 6D head movements with a motion-aware recurrent neural network (RNN). In this way, the predicted head poses act as the low-frequency holistic movements of a talking head, thus allowing our latter network to focus on detailed facial movement generation. To depict the entire image motions arising from audio, we exploit a keypoint based dense motion field representation. Then, we develop a motion field generator to produce the dense motion fields from input audio, head poses, and a reference image. As this keypoint based representation models the motions of facial regions, head, and backgrounds integrally, our method can better constrain the spatial and temporal consistency of the generated videos. Finally, an image generation network is employed to render photo-realistic talking-head videos from the estimated keypoint based motion fields and the input reference image. Extensive experiments demonstrate that our method produces videos with plausible head motions, synchronized facial expressions, and stable backgrounds and outperforms the state-of-the-art.

\end{abstract}

\begin{figure}[t]\centering
 \includegraphics[width=0.45\textwidth]{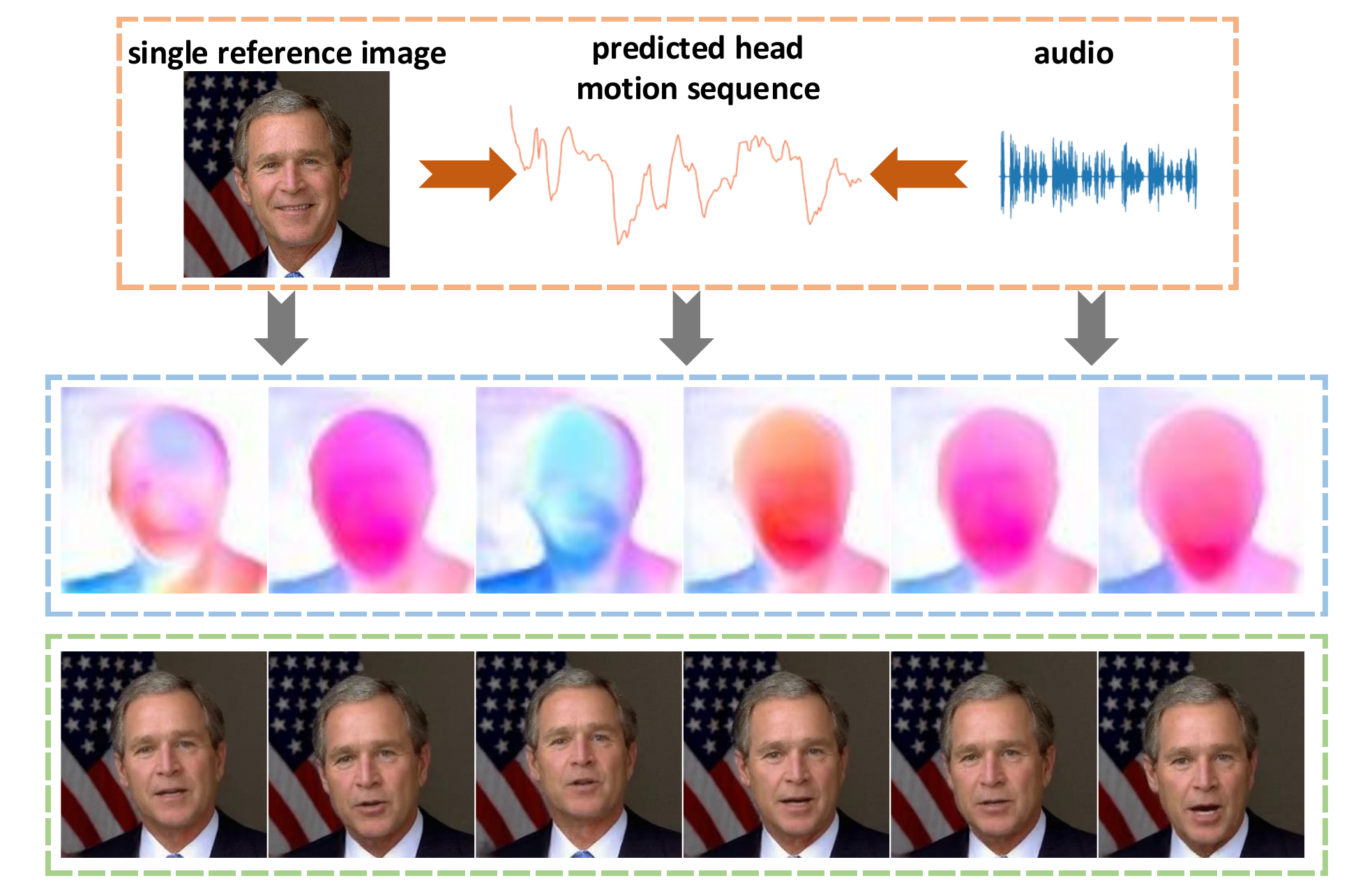}
 \caption{Illustration of the proposed audio-driven single image based talking-head video generation method. First row: the input reference image and audio, and the predicted head pose; middle row: generated motion fields from the audio and image; bottom row: synthesized talking-head frames.}
 \label{Fig: illustration}
\end{figure}

\section{Introduction}

Delivering information in an audio-visual manner is more attractive to humans compared to an audio-only fashion. Given an audio clip and one image of an arbitrary speaker, authentic audio-visual content creation has received great attention recently and also has widespread applications, such as human-machine interaction and virtual reality. 
A large number of one-shot talking-head works \cite{chen2019hierarchical,zhou2019talking,ijcai2019-129,vougioukas2019realistic,prajwal2020lip,zhu2020arbitrary} have been proposed to synchronize audio and lip movements. However, most of them neglect to infer head motions from audio, and a still head pose is less satisfactory for human observation. 
Natural and rhythmic head motions are also one of the key factors in generating authentic talking-head videos \cite{Ding2013HeadEyebrow,chen2020talking}.

\begin{figure}[t]\centering
 \includegraphics[width=0.48\textwidth]{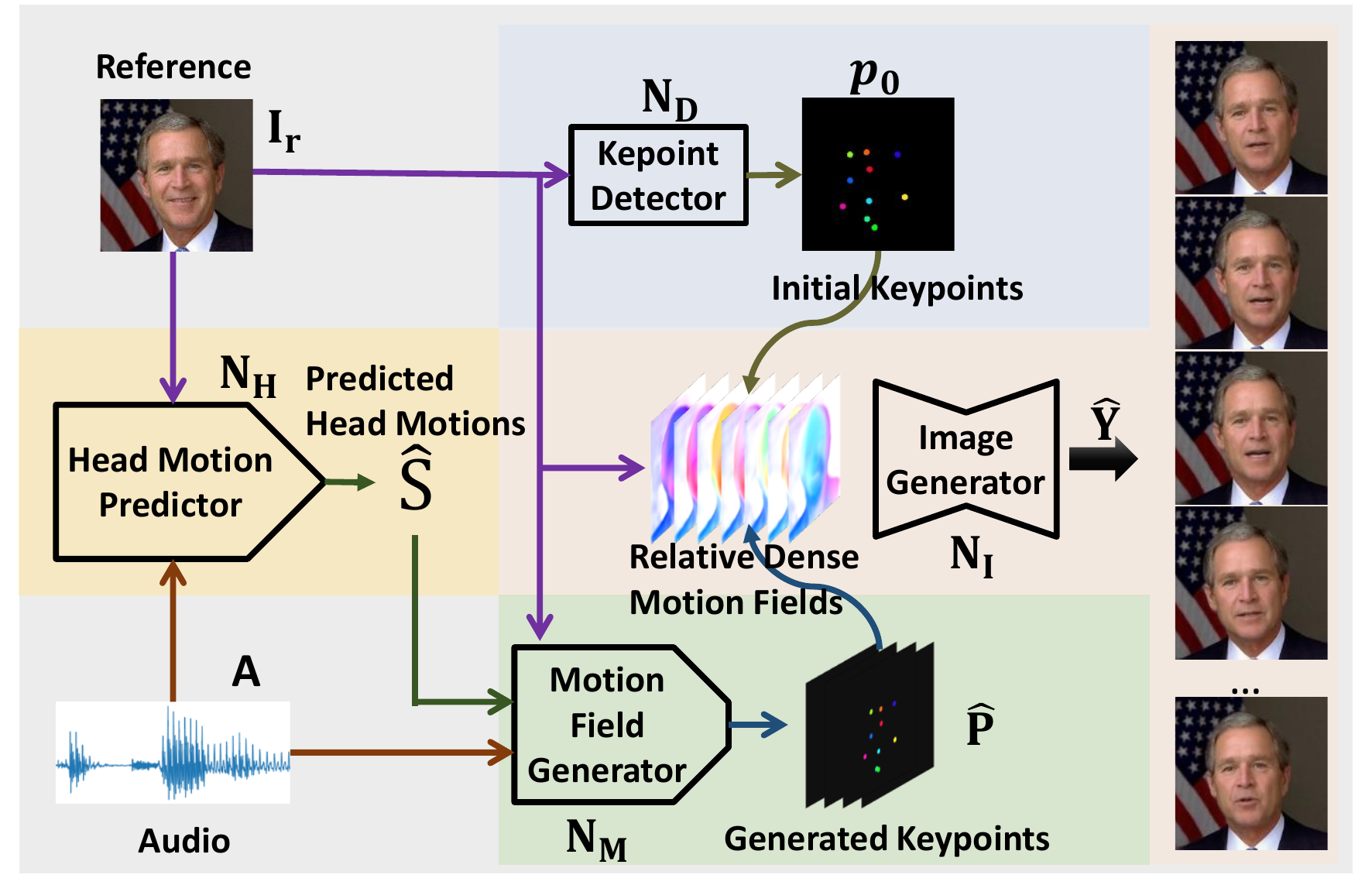}
 \caption{Pipeline of the proposed framework.}
 \label{Fig: pipeline}
\end{figure}

Albeit recent works \cite{chen2020talking,zhou2020makelttalk} take head movements into consideration, they often suffer from the ambiguous correspondences between head motions and audio in the training dataset, and fail to produce realistic head movements. 
For example, given the same audio content, one performer may move the head from left to right while another may move the head from right to left. These examples would introduce ambiguity in training and thus a network will produce a still-alike head to minimize the head motion loss.     
Moreover, when generating head motions, reducing artifacts in background regions is critical as well. However, the methods \cite{chen2020talking,zhou2020makelttalk} only focus on the face regions and thus result in obvious distortions in non-face regions (\eg, hair and background).
Therefore, how to produce realistic head motions and artifact-free frames for one-shot talking-head generation is still challenging. 


In this paper, a novel one-shot talking-head video generation framework is proposed to produce natural and rhythmic head motions while remarkably suppressing background artifacts caused by head motions.
Firstly, we propose to disentangle head motions and expression changes since head motions might exhibit ambiguity as aforementioned.
The head motion is modeled as a rigid 6 degrees of freedom (DoF) movement and it will act as the low-frequency holistic movement of talking head. We generate head motions via a motion aware recurrent neural network (RNN) (see Figure \ref{Fig: illustration}) and let another network focus on producing detailed facial movements.

Considering the ambiguous correspondences between head motions and audio, we opt to enforce the structural similarity between 
the generated and the ground truth head motion matrices making up of successive 6D motion vectors,
thus achieving more diverse head motions.

After obtaining head motions, we resort to a keypoint based dense motion field representation \cite{siarohin2019first} to depict the entire image content motions, including facial region, head and background movements. 
We develop an image motion field generator to produce keypoint based dense motion fields from the input audio, head poses and reference image, which will be used to synthesize new frames.
The relative dense motion field is then described by the differences between the predicted keypoints and those of the reference image (see Figure \ref{Fig: pipeline}).
Compared to the 3D face model or landmark-based representations that are used in prior works, the keypoint based representation models the motions of the facial regions, head, and backgrounds integrally, allowing for better governing the spatial and temporal consistency in the generated videos.
Finally, an image generation network~\cite{yu2018face,yu2018imagining,yu2019semantic,yu2019can} is employed to render photo-realistic talking-head videos from the estimated keypoint based motion fields and the input reference image.

\begin{figure}\centering
 \includegraphics[width=0.45\textwidth]{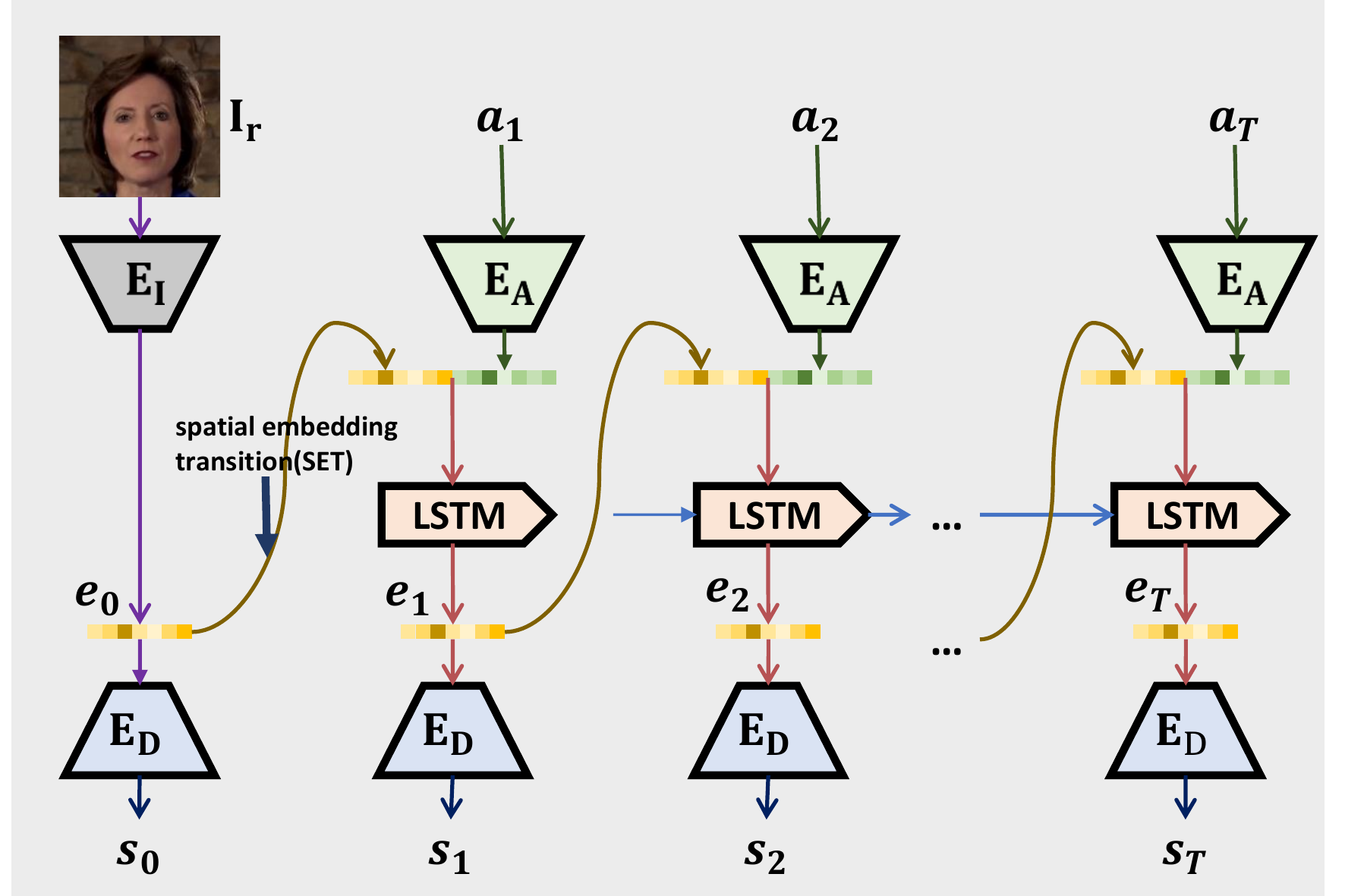}
 \caption{Architecture of the head motion predictor.}
 \label{Fig: head_motion}
\end{figure}

Extensive comparisons with the state-of-the-art methods show that our method achieves superior visual quality and authentic head motions without introducing noticeable artifacts.
In summary, we make the following technical contributions:
\begin{itemize}
\item We develop a new audio-driven talking-head generation framework that produces photo-realistic videos with natural head motions from a single image. 
\item We design a motion aware recurrent neural network to predict natural-looking head motions that match the input audio rhythm.
\item We present an image motion field generator to produce keypoint based dense motion fields and thus are able to govern the spatial and temporal consistency of the generated videos.
\item We achieve state-of-the-art results in terms of visual quality and rhythmic head motions.
\end{itemize}

\section{Related Work}
\subsection{Speech-driven Talking-head Generation}
Given the input audio, some approaches achieve great success to synthesize the talking head of a specific speaker \cite{suwajanakorn2017synthesizing,fried2019text,thies2020neural,li2021write}.
However, they need to be trained on minutes to hours of videos of each speaker.
Other works aim to reduce the speaker’s reference information to a single image.
Song \etal \shortcite{ijcai2019-129} and Vougioukas \etal \shortcite{vougioukas2019realistic} take the temporal dependency into account and generate the talking face video with GANs.
Zhou \etal \shortcite{zhou2019talking} design an end-to-end framework to generate videos from the disentangled audio-visual representation.
Prajwal \etal \shortcite{prajwal2020lip} employ a carefully-designed lip-sync discriminator for better lip-sync accuracy. 
These works directly learn the mapping from audio to facial pixels, which tend to produce blurry results.
For better facial details, some works utilize intermediate representations to bridge the variation of audio and pixels. 
Chen \etal \shortcite{chen2020talking} and Zhang \etal \shortcite{zhang2021flowguided} predict the coefficients of a 3D face model from audio, which is used to guide the image generation. 
Chen \etal \shortcite{chen2019hierarchical} and Zhou \etal \shortcite{zhou2020makelttalk} learn to generate facial landmarks from audio first, and then synthesize images from landmarks. 
Although these methods improve the visual quality of the inner face, 
none of the mediums models the non-face regions (e.g. hair and background). Hence, the texture and the temporal consistency of the outer face regions are not as good as that of the inner face.

\begin{figure}
\centering
\includegraphics[width=0.48\textwidth]{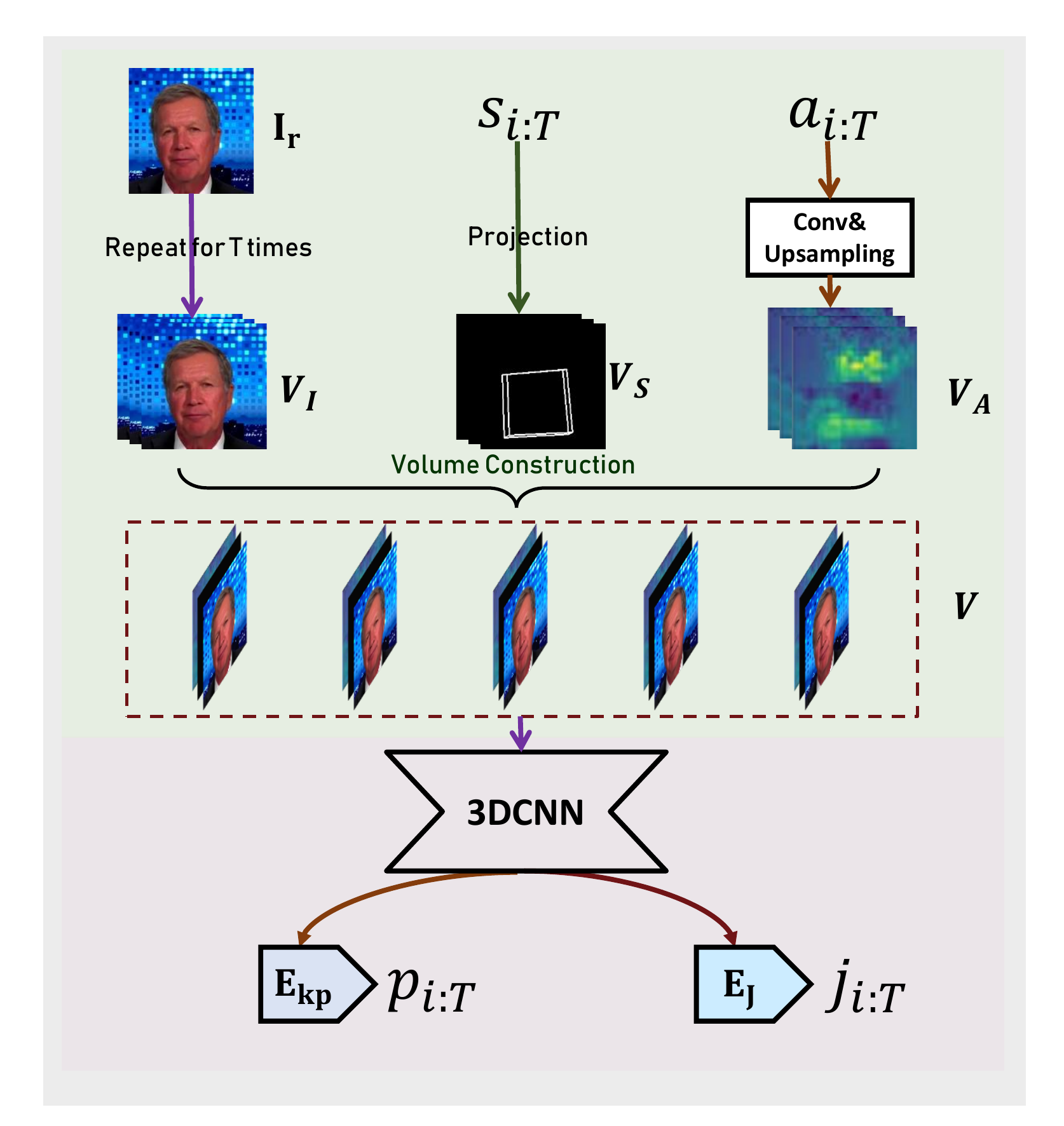}
\caption{Architecture of the motion field generator.}
\label{fig:keypoint generator}
\end{figure}

\subsection{Video-driven Talking-head Generation}

Video-driven methods control the motions of a subject with a driving video. 
Subject-specific methods \cite{bansal2018recycle,kim2018deep} focus on a specific person by training their models on that person. 
Subject-independent approaches drive the reenactment using landmarks \cite{Zhang2020CVPR,ha2020marionette,nirkin2019fsgan}, latent embeddings \cite{burkov2020neural} or feature warping \cite{wang2019fewshotvid2vid}.
Recently, Siarohin \etal \shortcite{siarohin2019first} represent the dense motion flow of the driving video as self-learned keypoints, and use the flow to warp the features of the reference image for reenactment. High visual quality is obtained due to the dense intermediate representation of the entire image.

\begin{figure}[t]\centering
 \includegraphics[width=0.48\textwidth]{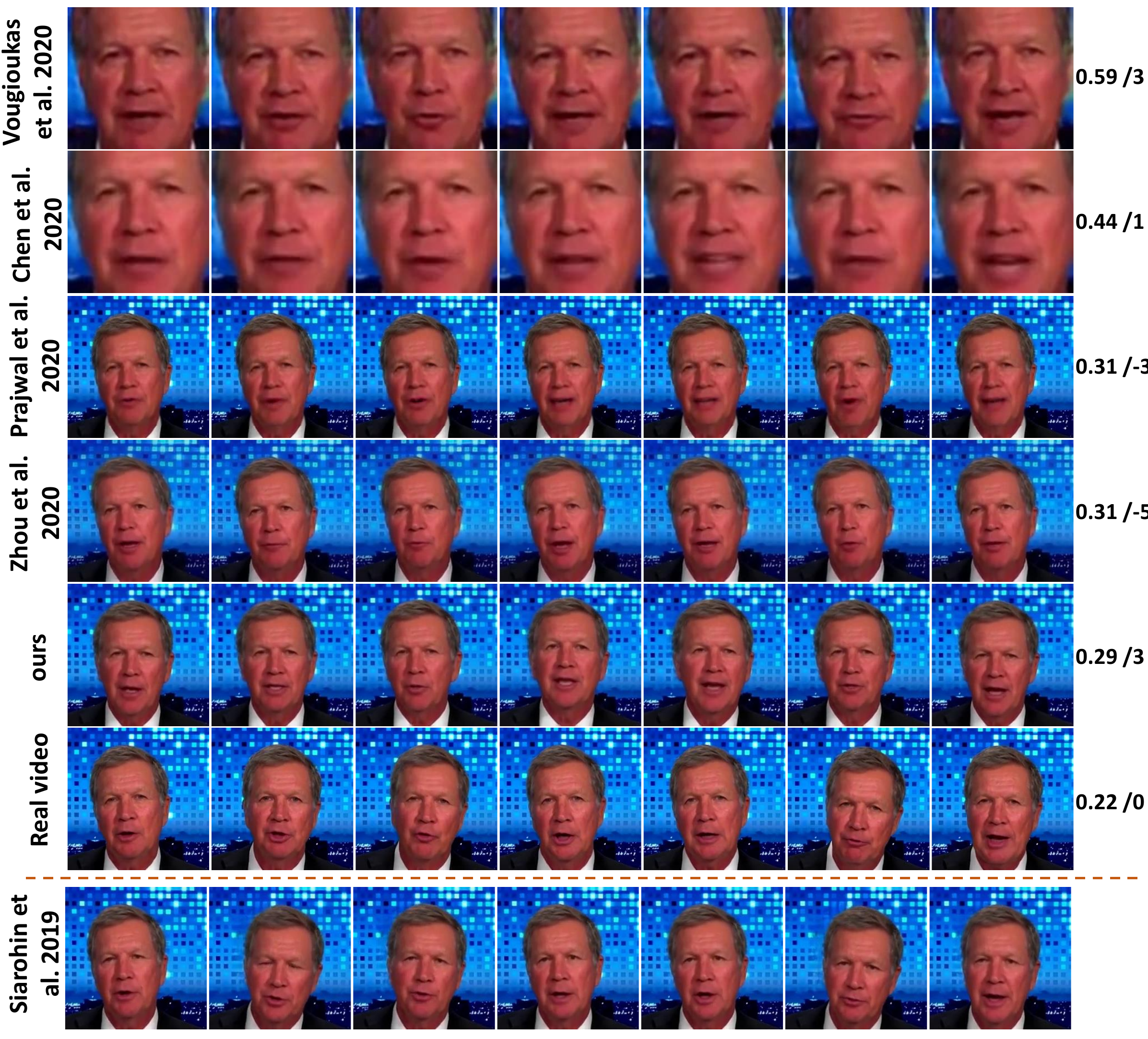}
 \caption{Comparison with the state-of-the-art. Please see more dynamic demos in our supplementary materials.}
 \label{Fig:qualitative}
\end{figure}

\subsection{Head Motion Prediction}

Traditional head motion prediction methods are designed for 3D avatars, which only involve the 3D head rotation \cite{Ding2013HeadEyebrow,greenwood2017predicting,sadoughi2018novel}. 
Methods designed for 2D images, on the other hand, need to produce both head rotation and translation.
Chen \etal \shortcite{chen2020talking} constrain the mean value and the standard deviation of the predicted head pose sequence to be similar to that of the real sequence. However, the statistical constraints cannot model the local motion details.
Zhou \etal \shortcite{zhou2020makelttalk} constrain the predicted facial landmarks to be consistent with that of the ground truth. Due to the L2 loss term, their method suffers from the ambiguous correspondences between head motion and audio, and converge to the slightly swinging. 
Different from the above two works, our method constrains the full head motion sequence for more natural head motion patterns. 


\section{Proposed Method}
\subsection{Overview}

Our method takes a reference image $\mathbf{I_r}$ and an audio clip $\mathbf{A}$ as input, and synthesizes video frames $\mathbf{\hat{Y}} = \hat{y}_{1:T}$ of the reference speaker synchronized with $\mathbf{A}$.
As illustrated in Figure \ref{Fig: pipeline}, the pipeline of our method consists of four components.

\paragraph{Head Motion Predictor $\headmotionpredictor$.} 
As the representation of low-frequency holistic movements,
the head poses are predicted individually. 
From both $\mathbf{I_r}$ and $\mathbf{A}$, $\headmotionpredictor$ produces the natural-looking and rhythmic head motion sequence $\mathbf{\hat{S}} = \hat{s}_{1:T}$.

\paragraph{Motion Field Generator $\keypointgenerator$.} 
$\mathbf{N_{M}}$ produces the self-learned keypoint sequence $\mathbf{\hat{P}} = \hat{p}_{1:T}$ that controls the dense motion field. 
$\mathbf{\hat{P}}$ contains the information of synchronized facial expressions, head motions, and non-face region motions.

\paragraph{Keypoint Detector $\keypointdetector$ and Image Generator $\imagegenerator$.} 
$\mathbf{N_{D}}$ detects the initial keypoints $p_0$ from $\mathbf{I_r}$. $\imagegenerator$ renders the synthesized images from the relative dense motion between $\mathbf{\hat{P}}$ and $p_0$.
The network architectures of $\keypointdetector$ and $\imagegenerator$ are adopted from \cite{siarohin2019first}.

\subsection{Head Motion Predictor}\label{headmotionpredictor} 

\begin{table}
    \centering
    
    \scalebox{0.85}{
    \begin{tabular}{ccccc}
    \toprule
    \multicolumn{2}{c}{} & PSNR $\uparrow$ & SSIM $\uparrow$ & FID$\downarrow$  \\
    \midrule
    \multirow{3}*{LRW} & Prajwal \etal \shortcite{prajwal2020lip}                     & \textbf{20.17} & 0.62 & \textbf{42.41} \\
    ~ & Zhou \etal \shortcite{zhou2020makelttalk}                    & 19.20 & 0.59 & 48.13  \\
    ~ & ours                                                         & 19.53 & \textbf{0.63} & 42.55  \\
    \midrule
    \multirow{3}*{GRID} & Prajwal \etal \shortcite{prajwal2020lip}                      & 28.82 & 0.86 & 31.14 \\
    ~ & Zhou \etal \shortcite{zhou2020makelttalk}                     & 24.55 & 0.78 & 34.57 \\
    ~ & ours                                                          & \textbf{30.93} & \textbf{0.91} & \textbf{22.50} \\ 

    \midrule
    \multirow{3}*{VoxCeleb} & Prajwal \etal \shortcite{prajwal2020lip}                          & 17.33  & 0.50  & 61.58  \\
    ~ & Zhou \etal \shortcite{zhou2020makelttalk}                         & 17.45  & 0.50  & 53.95  \\
    ~ & ours                                                              & \textbf{21.19}  & \textbf{0.68}  & \textbf{39.86}  \\    

    \bottomrule
    \end{tabular}
    }
    \caption{Quantitative comparison with the state-of-the-art.}
    \label{tab:Quantitative results.}
\end{table}

As preprocessing, the input raw audio $\mathbf{A}$ is first converted to an acoustic feature sequence $a_{1:T}$. $a_i$ refers to an acoustic feature frame. To be consistent with the frequency of the videos sampled at 25 fps, $a_i \in \mathbb{R}^{4 \times 41}$ makes up of acoustic features from $4$ successive sliding windows. In total, $41$ acoustic features are extracted from each sliding window, including 13 Mel Frequency Cepstrum Coefficients (MFCC), 26 Mel-filterbank energy features (FBANK), pitch and voiceless. The sliding window has a window size of 25ms and a step size of 10ms. 



In our work, $a_{1:T}$ is used to predict a head pose sequence describing head rotation and translation in the camera coordinate system. For reference images with different head scales and body positions, the natural head trajectories in pixel coordinates are also different. Specially, we employ an encoder $\mathbf{E_I}$ (ResNet-34) to extract the initial head and body state from $\mathbf{I_r}$. The extracted spatial embedding $e_0$ encodes the initial head rotation and translation. Afterwards, we use a two-layer LSTM to create natural head motion sequence that matches audio rhythm, as shown in Figure \ref{Fig: head_motion}. At each time step $i$, we first extract the audio embedding with another ResNet-34 encoder $\mathbf{E_A}$ from $a_i$ and concatenate it with the spacial embedding $e_{i-1}$ at step $i-1$. Then, the LSTM takes the concatenated embeddings as input and outputs the current $e_i$. Such spatial embedding transition (SET) passes the previous $e_{i-1}$ to the next time step, and therefore contribute to more natural head motions and better synchronization with audio.
Finally, a decoder $\mathbf{E_D}$ is used to decode $e_i$ to head pose $\hat{s}_i \in \mathbb{R}^6$ (3 for rotation and 3 for translation).  
Our head motion predictor supports an arbitrary length of audio input. The procedure is formulated as:
\begin{gather}
    (h_i,e_i) = \mathbf{LSTM}(h_{i-1},\mathbf{E_A}(a_i) \oplus e_{i-1}), \\
    \hat{s_i} = \mathbf{E_D}(e_i),
\end{gather}
where $h_i$ is the hidden state of step $i$, $\oplus$ means concatenation.

Since the mapping from audio to head motion is one-to-many mapping, the widely-used L1 loss and L2 loss are not suitable choices. Instead, we treat $\hat{s}_{0:T} \in \mathbb{R}^{6 \times T}$ as an image of size $6 \times T$, and impose the structural constraint on it using the Structural Similarity (SSIM)\cite{wang2004image} loss:
\begin{equation}
    \mathcal{L}_{SSIM} =  1 - \frac{(2 \mu \hat{\mu} + C_1) (2 {cov}+C_2))}{(\mu^{2} + \hat{\mu}^{2} + C_1)(\sigma^{2} + \hat{\sigma}^{2} + C_2))}.
\end{equation}
$\hat{\mu}$ and $\hat{\sigma}$ are the mean and standard deviation of $\hat{s}_{0:T}$, and ${\mu}$ and ${\sigma}$ are that of the groundtruth head pose sequence extracted by OpenFace \cite{baltrusaitis2018openface}. $cov$ is the covariance. $C_1$ and $C_2$ are two small constants. 
To improve the fidelity and smoothness of the predicted head motions, we also employ a discriminator $\mathbf{D}$ based on the PatchGAN. Specifically, we adapt the original PatchGAN to perform 1D convolution operations on head motion sequence along temporal trunks. 
The total loss function for $\headmotionpredictor$ is defined by:
\begin{equation}
    \mathcal{L}_\headmotionpredictor =  arg \min\limits_\headmotionpredictor \max \limits_\mathbf{D} (\headmotionpredictor,\mathbf{D}) +  \mathcal{L}_{SSIM}.
\end{equation}
The GAN loss is calculated by LSGAN. When training $\mathbf{N_H}$, we set the window length T to $256$.

\subsection{Motion Field Generator}\label{keypointgenerator}

From the $\mathbf{I_r}$, $a_{1:T}$ and the predicted $s_{1:T}$, the motion field generator produces the keypoints sequence that controls the dense motion field.
For the $i$-th frame, the keypoints include N positions $\hat{p}_i \in \mathbb{R}^{N\times2}$, and the corresponding Jacobian $\hat {j}_i \in \mathbb{R}^{N\times2\times2}$. 
Each $(\hat {p}_i^k,\hat {j}_i^k)$ pair represents a local image affine transform. 
The N local affine transforms with adaptive masks constitute the dense motion field \cite{siarohin2019first}.

Figure \ref{fig:keypoint generator} shows the structure of $\keypointgenerator$.
The multimodal input $s_{1:T}$, $\mathbf{I_r}$ and $a_{1:T}$ are first converted to a unified structure. 
$\mathbf{I_r}$ is expected to provide the identity constraint for generated keypoints, we downsample $\mathbf{I_r}$ to a size of $[W, H]$ and repeat it $T$ times as the tensor $V_I$ of size $[W,H,3,T]$.
Instead of directly taking $s_{1:T}$ as input, we draw a 3D box in the camera coordinate system to represent the head pose and project it to the image. In this way, we render a binary image for each pose frame, and stack them to get the pose tensor $V_S$ of size $[W,H,1,T]$.
As to $a_{1:T}$, we encode each $a_i$ to the feature map of the shape of $[W,H,2]$ using an encoder composed with conv and upsampling operations. The feature maps are also stacked as the tensor $V_A$ with size $[W,H,2,T]$.
Finally, we construct $V$ by concatenating $V_I,V_S,V_A$ along the channel dimension. In our experiments, $W$ and $H$ are set to 64, $T$ is set to 64.
Afterwards, we employ the 3D Hourglass Network (Hourglass-3D) to deal with the temporal dependence and ensure the smoothness of motion field between consecutive frames.
The Hourglass-3D takes $V$ as input and outputs a group of continuous latent features, which are then decoded into $\hat{p}_{1:T}$ and $\hat {j}_{1:T}$ by two decoders $\mathbf{E_{kp}}$ and $\mathbf{E_J}$ separately.

\begin{figure}[t]\centering
 \includegraphics[width=0.48\textwidth]{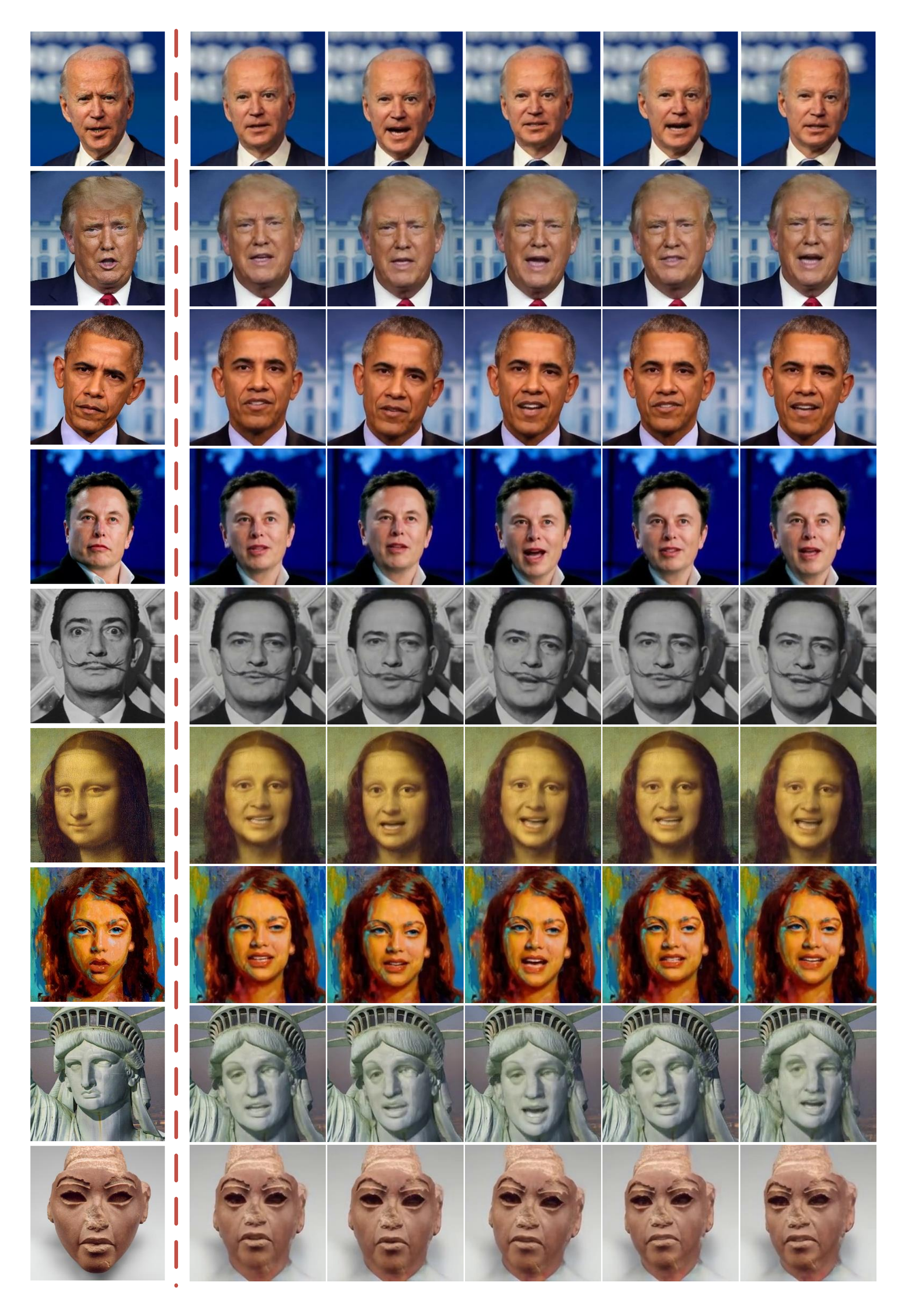}
 \caption{Samples generated with the same audio. Please zoom in for more details.}
 \label{Fig: moreresults}
\end{figure}

\begin{figure}[t]\centering
 \includegraphics[width=0.48\textwidth]{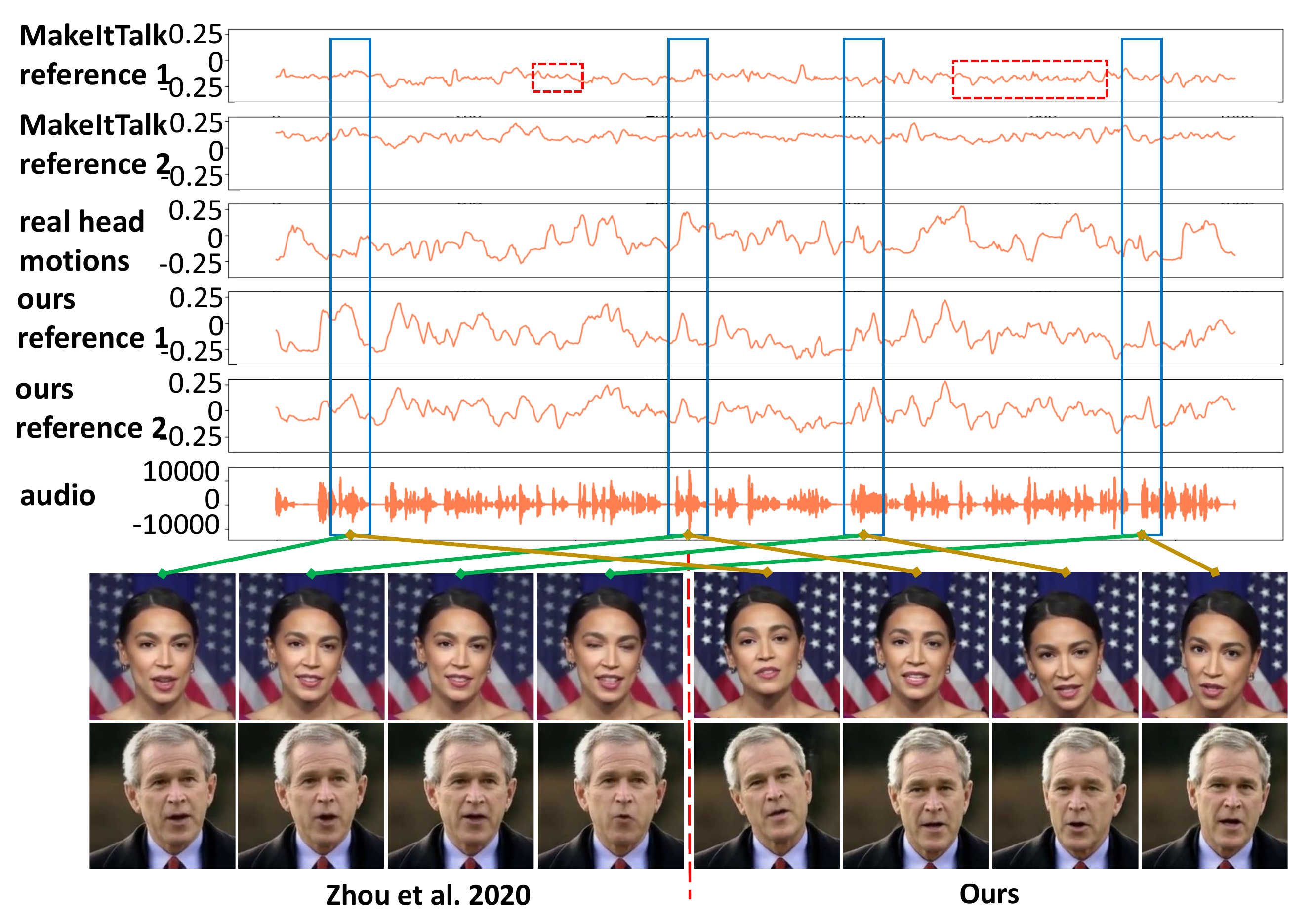}
 \caption{Comparison of head motion patterns on two reference images. Frames in the bottom are sampled from the blue boxes.}
 \label{Fig: headpose}
\end{figure}

The training process of $\keypointgenerator$ goes through two stages.
In the first stage, we use the pretrained $\mathbf{N_D}$ as guidance. 
The loss function is defined as:
\begin{align}
    \mathcal{L}_\mathbf{N_M^{1}} = \frac{1}{T} \sum_{i=1}^T(\lambda_m\mathbf{L_1}(\hat{m}_i,m_i)+\lambda_p\mathbf{L_1}(\hat{p}_i,p_i)+ \nonumber \\
    +\lambda_j\mathbf{L_1}(\hat{j}_i,j_i)).
\end{align}
$\mathbf{L_1}(\cdot,\cdot) $ denotes the L1 loss. $p_i$ and $j_i$ are the positions and Jacobians extracted by the pretrained $\keypointdetector$ from the training video. 
$\hat{m}_i$ and $m_i$ are heatmaps produced by $\mathbf{E_{kp}}$ and $\keypointdetector$, the keypoint positions are estimated from these heatmaps as in \cite{siarohin2019animating}. The heatmap term helps the convergence in the beginning.
$\lambda_p$ and $\lambda_j$ are both set to 10. $\lambda_m$ is set to 1 and decays to zero after a certain time. 

In the second stage, we import the pretrained $\mathbf{N_I}$ to help the fine-tuning of $\keypointgenerator$. 
Giving the predicted $\hat{p_i}$, $\imagegenerator$ renders the reconstructed frame $\hat{I_i}$ for the reconstruction loss:
\begin{equation}
    \mathcal{L}_{rec}(\hat{I},I) = \sum_{i=1}^l \mathbf{L_1}(C_i(\hat{I}),C_i(I)),
\end{equation}
where $C_i(\cdot)$ is the $i$th channel feature of a specific pretrained VGG-16 layer with $l$ channels. 
We apply $\mathcal{L}_{rec}(\hat{I},I)$ on the image pyramid of multiple resolutions, and sum them as $\mathcal{L}_{rec}^{mul}(\hat{I},I)$.
For more stable performance, we also adopt the equivariance constraint losses of \cite{siarohin2019first}, denoted as $\mathcal{L}_{eq}^{P}(\hat{p})$ and $\mathcal{L}_{eq}^{J}(\hat{j})$ respectively. Please refer to \cite{siarohin2019first} for more details. 
The total loss function is defined as:
\begin{align}
\mathcal{L}_\mathbf{N_M^{2}} = \frac{1}{T} \sum_{i=1}^T(\lambda_p'\mathbf{L_1}(\hat{p}_i,p_i)+\lambda_{rec}\mathcal{L}_{rec}^{mul}(\hat{I}_i,I_i)+ \nonumber \\
+ \lambda_{eq}^P \mathcal{L}_{eq}^{P}(\hat{p}_i) + \lambda_{eq}^J \mathcal{L}_{eq}^{J}(\hat{j}_i).
\end{align}
$\lambda_p'$ is set to 100, $\lambda_{rec}$, $\lambda_{eq}^P$ and $\lambda_{eq}^J$ are all set to 10.

\section{Experiment Setup}

\subsection{Datasets}
We use prevalent benchmark datasets \textbf{VoxCeleb} \cite{nagrani2017voxceleb}, \textbf{GRID} \cite{cooke2006audio} and \textbf{LRW} \cite{chung2016lip}
to evaluate the proposed method. 
\textbf{VoxCeleb} consists of speech clips collected from YouTube.
\textbf{GRID} contains video clips of 33 speakers in the experimental condition. \textbf{LRW} contains 500 different words spoken by hundreds of people in the wild.
Specially, for \textbf{VoxCeleb} and \textbf{GRID}, we re-crop and resize the original videos to $256\times256$ as in \cite{siarohin2019first}, to get 54354 and 25788 shot video clips respectively. For \lrw, we use its original format by aligning the face in the middle of each frame.
We split each dataset into training and testing sets following the setting of previous works. All videos are sampled at 25 fps.

\subsection{Implementation Details}

All our networks are implemented using PyTorch. We adopt Adam optimizer during training, with an initial learning rate of 2e-4 and weight decay to 2e-6. 
$\headmotionpredictor$ is trained on VoxCeleb for one day on one RTX 2080 Ti with batchsize 64.
When training $\keypointdetector$, $\imagegenerator$ and $\keypointgenerator$, we separate \textbf{LRW} and the other two datasets into two groups due to the different cropping strategies. 
For each group, We first follow the training details of \cite{siarohin2019first} to train $\keypointdetector$ and $\imagegenerator$, then we train $\keypointgenerator$ with frozen $\keypointdetector$ and $\imagegenerator$.
The training of $\keypointdetector$ and $\imagegenerator$ takes 3 days with batchsize 28, and that of  $\keypointgenerator$ takes one week with batchsize 4 on 4 RTX 2080 Ti. Specially, for \textbf{LRW}, the window length is set to 32 because of the short video clips.




\section{Experiments Results}

\subsection{Evaluation of Visual Quality}

We show the comparisons with the state-of-the-art methods in Figure \ref{Fig:qualitative}, including Vougioukas \etal \shortcite{vougioukas2019realistic}, Chen \etal \shortcite{chen2019hierarchical}, Prajwal \etal \shortcite{prajwal2020lip} and Zhou \etal \shortcite{zhou2020makelttalk}. The samples are generated with the same reference image and audio. 
Instead of only synthesizing a fixed head or cropped face, our method creates more realistic videos with head motions and full background. Compared to Zhou \etal \shortcite{zhou2020makelttalk}, our results produce more plausible head movements and a more stable background with fewer artifacts. Besides, our method holds the identity of the speaker well even after a large pose change, owing to the motion field representation. 
The last row of Figure \ref{Fig:qualitative} shows the video-driven result of Siarohin \etal \shortcite{siarohin2019first}. Although our results are generated from audio instead of video, we show comparable results in visual quality.
We present more results in Figure \ref{Fig: moreresults} for unseen identities, including non-realistic paintings and human-like statues. The results show the excellent generalization ability of our method.

The quality of generated videos is evaluated using common reconstruction metrics SSIM, PSNR and Frechet Inception Distance (FID). We compare our approach with recent state-of-the-art methods including  Zhou \etal \shortcite{zhou2020makelttalk}, and Prajwal \etal \shortcite{prajwal2020lip}, which also synthesize images with background rather than cropped face only. The quantitative results are shown in Table \ref{tab:Quantitative results.}.
Comparing with Zhou \etal \shortcite{zhou2020makelttalk} and Prajwal \etal \shortcite{prajwal2020lip}, we achieve the highest PSNR/SSIM and the lowest FID score on \textbf{GRID} and \textbf{VoxCeleb}. 
Since videos of \textbf{LRW} are very short (about 1.16s), most of their head pose and background barely move. Prajwal \etal \shortcite{prajwal2020lip} only edits the mouth region of the reference image, resulting in the best PSNR and FID in \textbf{LRW}.

\subsection{Evaluation of Lip-sync}
We employ SyncNet \cite{chung2016lip} to evaluate the audio-visual synchronization of the proposed method. The metric values (confidence/offset) of each method are listed on the right side of Figure \ref{Fig:qualitative}. We obtain competitive lip-sync accuracy comparing with the state-of-the-art methods, even though we address a more challenging task.

\subsection{Evaluation of Head Motion}

Figures \ref{Fig: moreresults} shows that our head motion predictor creates natural and rhythmic head motions depending on both the audio and the identity.  
We further compare the head motion predictor with Zhou \etal \shortcite{zhou2020makelttalk} (MakeItTalk) with the same audio and two reference images. The head movements of MakeItTalk are detected from the generated videos. For better visualization, we reduce the six dimensional head motions into one dimension by PCA, and show the sequential results in Figure \ref{Fig: headpose}. 
MakeItTalk hardly changes the head orientation and contains many repetitive behaviors, as shown in the red boxes. Their head motion patterns tend to be slightly swinging around the initial pose. 
In contrast, our head motions preserve the rhythm and synchronization with audio, and are much closer to the ground truth, as shown in the blue boxes. 
Furthermore, we produce corresponding motion sequences with different input identities.


\begin{figure}[t]\centering
 \includegraphics[width=0.48\textwidth]{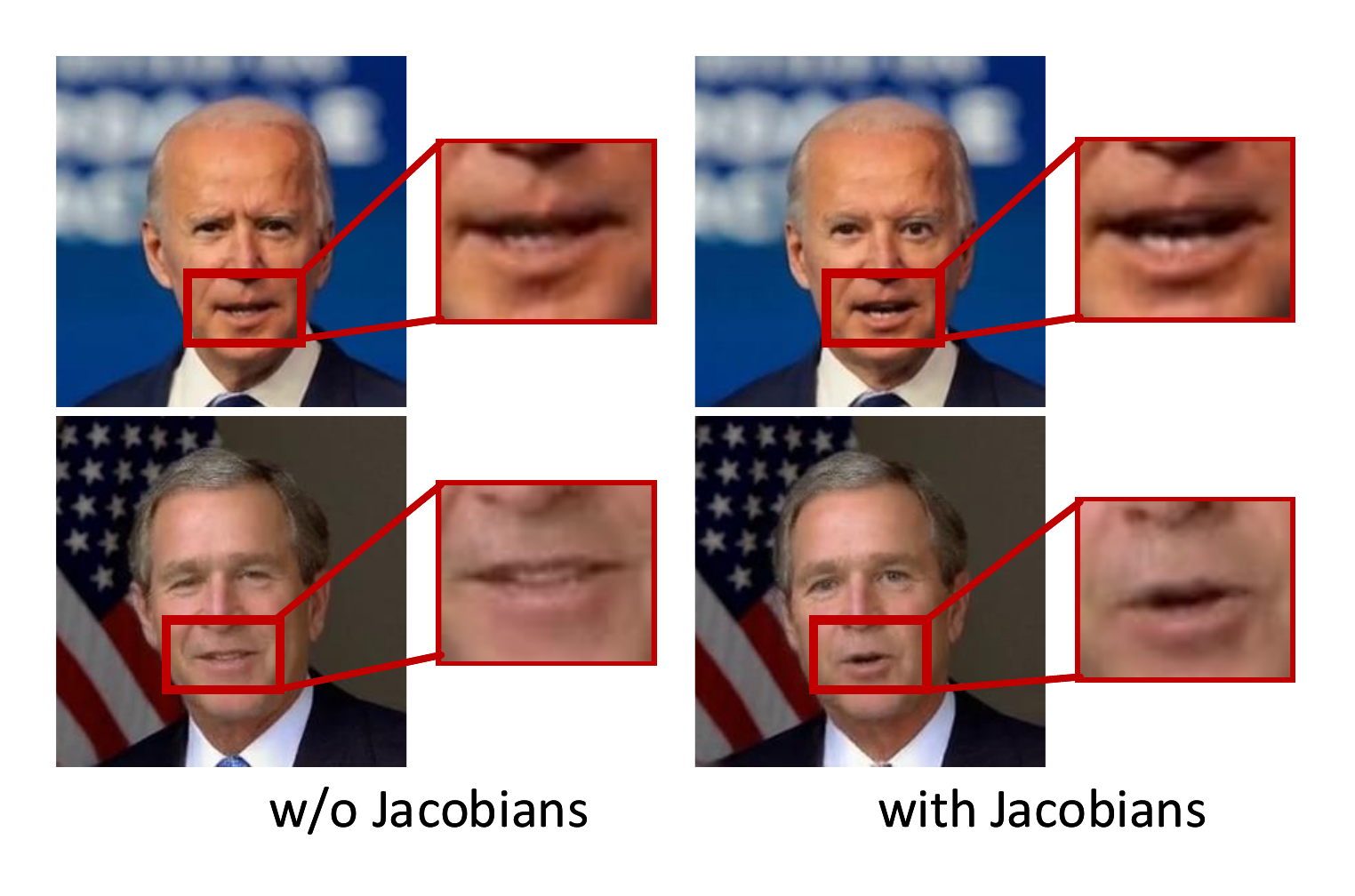}
 \caption{Results with and w/o Jacobians.}
 \label{Fig: ablation_jaco}
\end{figure}

\begin{figure}[t]\centering
 \includegraphics[width=0.48\textwidth]{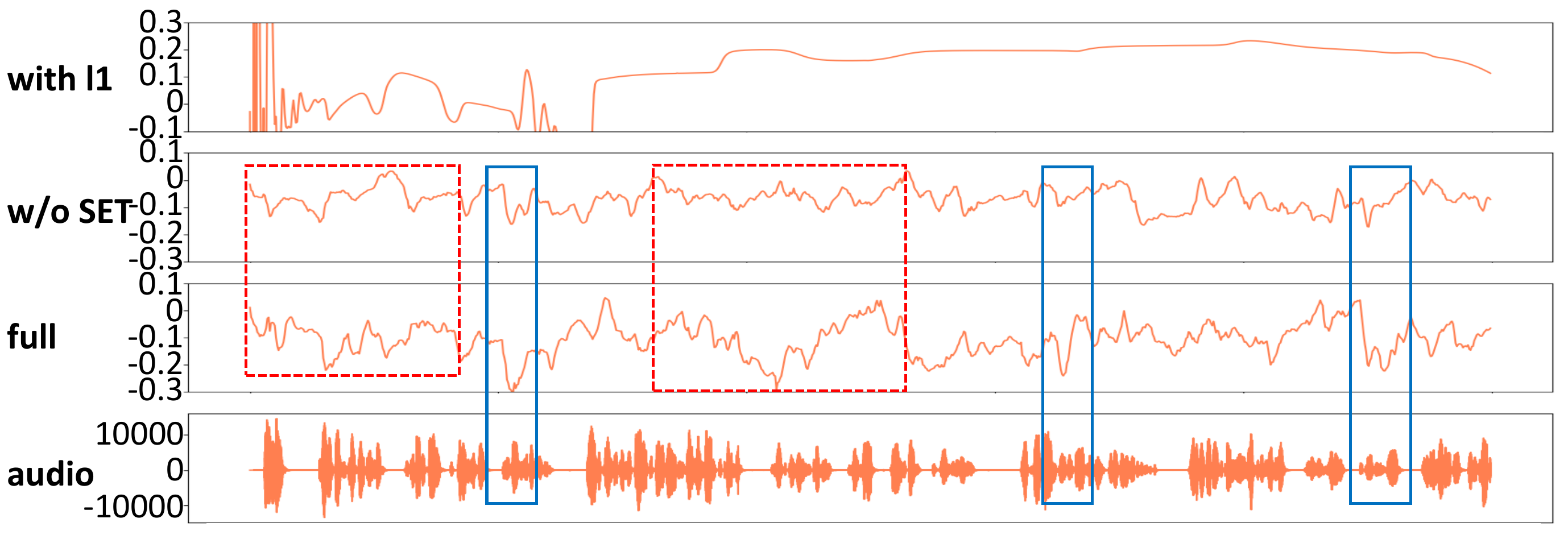}
 \caption{Ablation study on the head motion predictor.}
 \label{Fig: ablation_pose}
\end{figure}

\begin{figure}[t]\centering
 \includegraphics[width=0.48\textwidth]{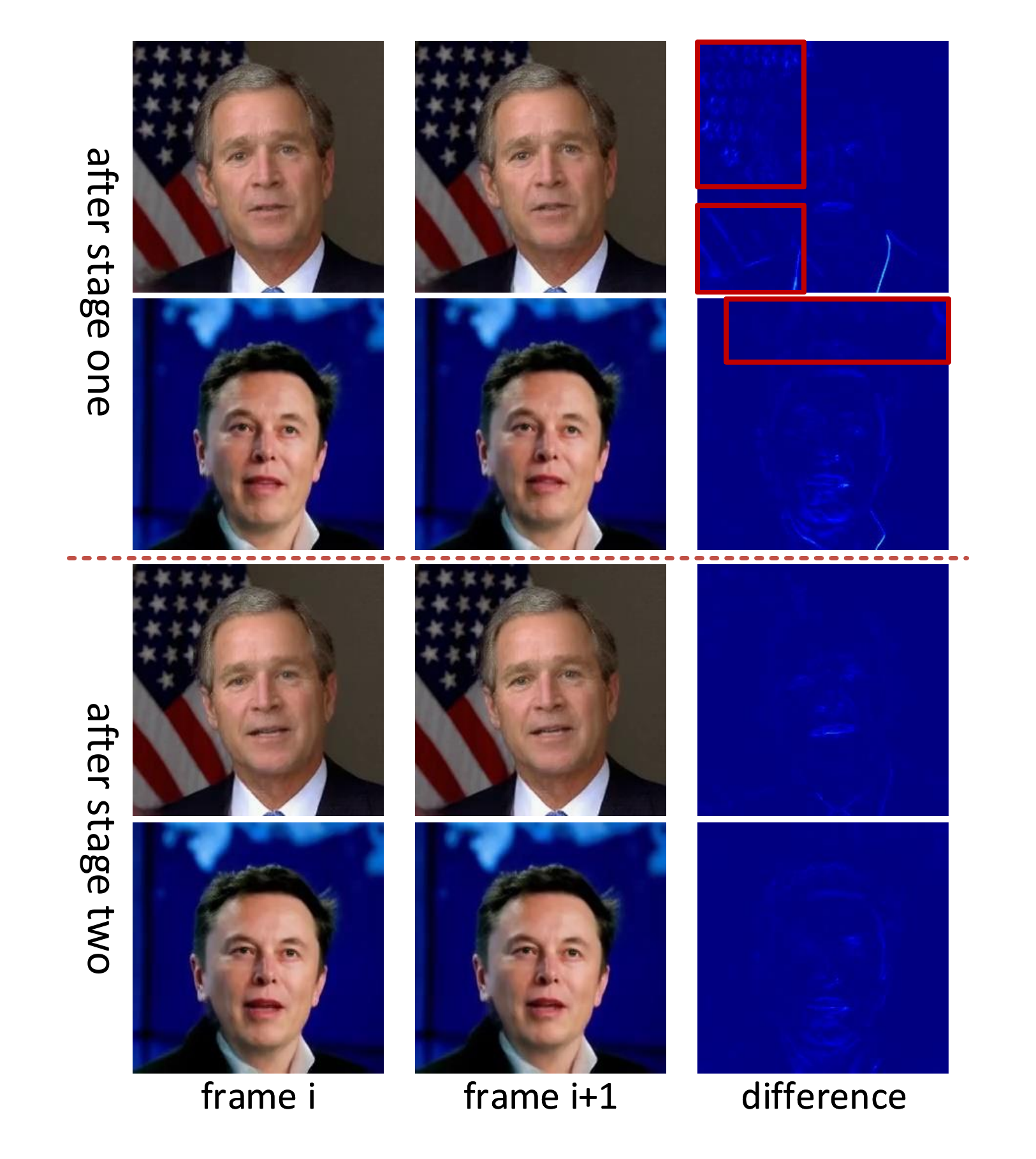}
 \caption{Results with one/two stage training. The red box shows the inconsistency between adjacent frames. Please zoom in for more details.}
 \label{Fig: ablation_stage}
\end{figure}

\subsection{Ablation Study}

We perform the quantitative evaluation of ablation study on VoxCeleb to illustrate the contribution of each component, the results are shown in Table \ref{tab:quantitative_results} . We construct three variants including GT-keypoint (using GT keypoints), GT-head (using GT head movements) and our full model. Here, we also evaluate L1 distance between predicted and GT head movement vectors, marked as HE, L1 distance between generated and GT keypoints marked as KE. Although there exists numerical differences, the generated videos are still natural-looking.

\begin{table}[h]
    \centering
    \footnotesize
    \begin{tabular}{c c c c }
    \toprule
    
             &  GT-keypoint  & GT-head  & full model   \\
    \midrule
    HE $\downarrow$ & - & - & 0.1910 \\
    KE $\downarrow$ & - & 0.0143 & 0.0630 \\ 
    PSNR $\uparrow$ & 26.40 & 25.37  &  21.19   \\ 
    SSIM $\uparrow$ & 0.82 & 0.79   & 0.68 \\ 
    \bottomrule
    \end{tabular}
    \caption{Quantitative results of ablation study on VoxCeleb.}
    \label{tab:quantitative_results}

\end{table}

We evaluate the effectiveness of Jacobians by removing $\mathbf{E_J}$ from $\keypointgenerator$. The generated result with and without Jacobians are shown in Figure \ref{Fig: ablation_jaco}. 
Without Jacobians, the lip seems to have only open-and-close patterns.
It illustrates that the local affine transformations benefit to the lip shape details. 


To show the effectiveness of the SSIM loss and SET in $\headmotionpredictor$, we conduct two variants by replacing the SSIM loss with L1 loss (with L1), or removing SET from our model (w/o SET).
The results are compared with our full method (full) in Figure \ref{Fig: ablation_pose}. 
The model trained with L1 loss creates unnatural head motion sequence, because it suffers from the one-to-many mapping ambiguity. 
Results of the model without SET contains less dynamics and is not well synchronized with audio, especially in the red and blue boxes.


We then compare the results with or without the second training stage in Sec.\ref{keypointgenerator}. We visualize the difference between two consecutive frames in Figure \ref{Fig: ablation_stage}. Results without the refinement of the second stage contain texture inconsistency and slight jitters. 
The constraint on pixel level is helpful for improving the temporal coherence and the fidelity of videos.


\subsection{User Study}


We further conduct an online user study to compare the integrated quality of our method with state-of-the-arts. We create 4 videos for each method with the same input, to obtain $4\times 5 = 20$ video clips. 33 participants are asked to rate "does the video look natural?" for each video from 1 to 5. The statistical results are shown in Table \ref{tab:userstudy}. 
Our method outperforms all compared methods significantly with the $66.7\%$ of the cases that are judged as natural. It indicates that in addition to lip-sync, people are also quite sensitive to both frozen head pose and background artifacts.
Besides, videos of only cropped faces \cite{chen2019hierarchical,vougioukas2019realistic} are rated lower compared with others.

\section{Conclusion and Discussion}

In this paper, we propose a novel framework for one-shot talking-head generation from audio, which creates high fidelity videos with natural-looking and rhythmic head motions. 
We decouple the head motions from full-frame audio-dependent motions and predict the head motions individually in accordance with audio dynamics. 
Then, the motion field generator produces the keypoints that control the dense motion field from audio and head poses. 
Finally, an image rendering network synthesizes the videos using the dense motion field. 
Our method is evaluated qualitatively and quantitatively.
The evaluation results show that our method predicts natural head motions, and produces few artifacts in non-face regions and between consecutive frames even though the head goes through a large pose change.
Our method is proved to have a higher visual quality compared to the state-of-the-art.


\begin{table}
    \centering
    \setlength{\tabcolsep}{1mm}{
    \scalebox{0.85}{
    \begin{tabular}{c c c c c c c c}
    \toprule
         & 1   &   2   & 3   & 4   & 5  & 'natural'(4+5) \\
    \midrule
         
Chen \etal \shortcite{chen2019hierarchical} & 8\%   &31\%   &39\%   &17\%  &5\%  &   22.0\%  \\ 
Vougioukas \etal \shortcite{vougioukas2019realistic}  & 41\%  &30\%  &14\%  &12\%  &2\%  &  14.4\%  \\ 
Prajwal \etal \shortcite{prajwal2020lip}      & 5\%  &20\%  &40\%  &30\%  &6\%   & 35.6\%  \\ 
Zhou \etal \shortcite{zhou2020makelttalk} & 8\%   &35\%  &30\%  &21\%  &7\%    & 28.0\%  \\ 
Ours     & 3\%   &8\%  &22\%  &39\%  &27\%   & \textbf{66.7\%}  \\  

    \bottomrule
    \end{tabular}
    }}
    \caption{Statistics of user study.}
    \label{tab:userstudy}
\end{table}

Although our method outperforms previous works, 
our lip-sync accuracy drops on the bilabial and labiodental phonemes such as \textit{p}, \textit{f} and \textit{m}. Compared to methods that focus on lip-sync, our framework trades a slight drop of lip-sync accuracy for much better head motion and visual quality. Such a trade-off is proved to be favored by most participants in our user study. 
We will be devoted to increasing the lip-sync accuracy without decreasing the current visual quality in future works.
Besides, our method cannot capture the blink pattern, and fails on input reference images with extreme pose or expressions, which also needs to be addressed in the future.


\section*{Ethical Impact}

With the convenience of creating photo-realistic videos for arbitrary identity and audio, our method has widespread positive applications, such as video conferencing and movie-dubbing. On the other hand, it may be misused by immoralists.
To ensure proper use, we will release our code and models to promote the progress in detecting fake videos.
Besides, we strongly require that any result created using our code and models must be marked as synthetic.


\bibliographystyle{named}
\bibliography{ijcai21}

\end{document}